\documentclass[11pt, onecolumn]{article}
\usepackage[top=1in, bottom=1in, left=1.25in, right=1.25in]{geometry}

\usepackage{amsfonts}
\usepackage[cmex10]{amsmath}
\usepackage{amsthm}
\usepackage{cite}
\usepackage{color,soul}
\usepackage{graphicx}
\usepackage{bm}
\usepackage{booktabs}
\usepackage{algorithm}
\usepackage{algorithmic}
\usepackage{mdwmath}
\usepackage{mdwtab}
\usepackage{multirow}
\usepackage{diagbox}
\usepackage{url}

\usepackage{float}
\usepackage{subcaption}
\usepackage{wrapfig}
\usepackage[export]{adjustbox}

\graphicspath {{figures/}}

\theoremstyle{definition}
\newtheorem{definition}{Definition}

\theoremstyle{definition}

\theoremstyle{definition}

\theoremstyle{definition}

\theoremstyle{definition}

\theoremstyle{remark}

\begin{document}

\title{Active Orthogonal Matching Pursuit for Sparse Subspace Clustering}

\author{Yanxi Chen, Gen Li, and Yuantao Gu% <-this % stops a space
\thanks{The authors are with the Department of Electronic Engineering and Tsinghua National Laboratory for Information Science and Technology (TNList), Tsinghua University, Beijing 100084, China. 
The corresponding author of this work is Y. Gu  (E-mail:\,gyt@tsinghua.edu.cn).
}}

\date{submitted June 20, 2017, accepted August 9, 2017, for publication in \\ \emph{IEEE Signal Processing Letters}}

\maketitle

\begin{abstract}

Sparse Subspace Clustering (SSC) is a state-of-the-art method for clustering high-dimensional data points lying in a union of low-dimensional subspaces. 
However, while $\ell_1$ optimization-based SSC algorithms suffer from high computational complexity,
other variants of SSC, such as Orthogonal Matching Pursuit-based SSC (OMP-SSC), lose clustering accuracy in pursuit of improving time efficiency.
In this letter, we propose a novel Active OMP-SSC, which improves clustering accuracy of OMP-SSC by 
adaptively updating data points and randomly dropping data points in the OMP process, while still enjoying the low computational complexity of greedy pursuit algorithms.
We provide heuristic analysis of our approach, and explain how these two active steps achieve a better tradeoff between connectivity and separation. 
Numerical results on both synthetic data and real-world data validate our analyses and show the advantages of the proposed active algorithm.

\end{abstract}

\section{Introduction}
\label{sec:introduction}

In a big-data era, unsupervised learning plays a significant role in analyzing numerous unlabeled data.
In many applications, such as motion segmentation and face clustering, high-dimensional data samples are drawn approximately from a union of low dimensional subspaces.
Subspace clustering (SC) \cite{vidal2011subspace} refers to the problem of clustering these data points into their original subspaces, 
and various methods have been developed to solve this problem
\cite{vidal2011subspace,elhamifar2013sparse,liu2013robust,vidal2014low,lu2012robust,lu2013correlation,lu2013correntropy,heckel2015robust,zhang2013robust,park2014greedy,yang2015sparse,you2016oracle,peng2017constructing,hu2014smooth,li2017structured,peng2016feature}. 
Among them, Sparse Subspace Clustering (SSC), the first one to introduce sparse representation into SC problem, is a state-of-the-art method, with elegant formulation and theoretical guarantees to work under weak conditions, e.g., at the presence of noise \cite{wang2013noisy} and outliers \cite{soltanolkotabi2012geometric}.
SSC algorithm solves SC problem by first finding a sparse self-representation for each data point, and then applying spectral clustering \cite{von2007tutorial} to the similarity matrix. 
However, it is inefficient to solve an optimization problem for sparse representation (as the original SSC algorithm does) when analyzing large-scale data.
One way to speed up SSC is replacing $\ell_1$ optimization with greedy pursuit, e.g.,  Orthogonal Matching Pursuit (OMP) \cite{tropp2007signal}, for sparse self representation \cite{dyer2013greedy,you2016scalable,tschannen2016noisy}, which improves the time efficiency of SSC by several orders of magnitude for large-scale problem.
Another way is to compress data points (i.e., to reduce dimension) before applying SSC algorithm, leading to Compressed Subspace Clustering algorithm \cite{mao2014compressed,heckel2017dimensionality,wang2016theoretical},
whose performance is guaranteed by works in random projection and Restricted Isometry Property (RIP) of subspaces \cite{li2017distance,li2017restricted}.

One motivation of this letter is the drawbacks of existing SSC algorithms. 
Traditional $\ell_1$ optimization-based SSC ($\ell_1$-SSC) suffers from high computational complexity, while OMP-SSC loses clustering accuracy.
Our goal is to develop a fast SSC algorithm with high clustering accuracy.
We are also motivated by the insight that the final clustering accuracy depends on two major properties of the similarity matrix: \emph{Subspace Detection Property} (SDP) \cite{soltanolkotabi2012geometric}, i.e., ``no false connection'', among points from different subspaces, and \emph{connectivity} among points from the same subspace.
While there is rich literature on theoretical analysis of SDP, much fewer works concern about the connectivity problem \cite{nasihatkon2011graph}.
\cite{wang2016graph} introduces a post-processing procedure,
and \cite{wang2013provable} modifies the optimization target as a weighted sum of $\ell_1$ norm and nuclear norm of the representation matrix in order to tradeoff between separation and connectivity.

In this letter, we proposed a novel algorithm, active OMP-SSC (A-OMP-SSC), which introduces two active-style processes, including adaptively updating and randomly dropping data points.
We analyze how the proposed algorithm obtains a better tradeoff between connectivity and SDP, thus improving clustering accuracy while preserving advantages of greedy methods in time efficiency.
Numerical results validate our analyses and show the advantages of our algorithm.

\section{Preliminary}
\label{sec:problem_formulation}

Given data matrix $\mathbf{X}=\left[{\bf x}_1,{\bf x}_2,\cdots,{\bf x}_N\right] \in \mathbb{R}^{D \times N}$, where $D$ is data dimension and $N$ is total number of data points, 
the aim of SC is to cluster these data points into their original subspaces.
In the preprocessing, all points are $\ell_2$ normalized, i.e., $\|\mathbf{x}_i\|_2=1,  \forall i$.

\subsection{SSC and OMP-SSC}
\label{sub:sparse_subspace_clustering_}

The core idea behind SSC is that, each data point can be approximately represented as a sparse linear combination of other points, 
where the nonzero entries correspond to those from the same subspace. 
The first step of SSC is to find a sparse representation for each point by solving an optimization problem of
\begin{equation}
\min_{\mathbf{c}_i} { } \|\mathbf{c}_i\|_1 + \frac{\lambda}{2} \|\mathbf{x}_i - \mathbf{X} \mathbf{c}_i\|_2^2 \quad \text{ s.t. } \quad c_{ii} = 0,
\label{eq:noisy}
\end{equation}
where $\mathbf{c}_i$ denotes the representation coefficient vector for ${\bf x}_i$
and $\lambda$ is a balance parameter.
The second step is to apply spectral clustering to the similarity matrix 
$
\mathbf{A} = |\mathbf{C}|+|\mathbf{C}|^{\rm T}
$
and get the clustering labels.

OMP-SSC is a faster version of SSC. It finds the sparse representation by OMP instead.
Given data point $\mathbf{x}_i$ and dictionary composed of all other data points, OMP iteratively finds an atom that has the largest absolute inner product with residual, adds this atom to the neighbor set, projects $\mathbf{x}_i$ onto the span of its current neighbors, and updates residual, until the iteration number reaches a certain value or the norm of residual is small enough.

\subsection{Geometric Analysis}
\label{sub:geometric_analysis}

Most literature focuses theoretical analyses of SSC on SDP.

\begin{definition}
\cite{soltanolkotabi2012geometric} Subspace Detection Property (SDP) holds if and only if it holds that for all $i$, $\mathbf{c}_i$ is non-trivial (i.e., not all-zeros), and $c_{ij} \neq 0$ only when $\mathbf{x}_i$ and $\mathbf{x}_j$ lie in the same subspace.
\end{definition}

A geometric framework for theoretical analysis of SDP was first proposed in \cite{soltanolkotabi2012geometric},
and has been thoroughly studied since then.
In comparison, only a few works analyze the connectivity problem of SSC. It is shown in \cite{nasihatkon2011graph} that, when subspace dimension exceeds three, there is no guarantee that data points from the same subspace will form a connected component even in the noiseless case.

\section{Active OMP-SSC}
\label{sec:active_omp_for_ssc}

\subsection{Intuition}
\label{sub:intuition}

As mentioned before, the final clustering accuracy of SSC depends on both connectivity and SDP of the similarity matrix obtained by the self-representation step.
Actually, SDP holding true is not necessary for correct clustering results due to the robustness of spectral clustering. 
On the other hand, spectral clustering could fail if the connectivity of similarity matrix is so weak that data points from the same subspace are separated.
The ideal case is that both connectivity and SDP are strong, which is unrealistic in practice at the presence of noise.
Therefore, the problem of increasing clustering accuracy becomes to tradeoff between connectivity and SDP. 

\subsection{Algorithm}
\label{sub:algorithm}

The proposed A-OMP-SSC algorithm operates iteratively on all data points.
For ${\bf x}_i$, we first find its sparse representation ${\bf c}_i$ on the current dictionary, indexed by a set $\mathcal{D}$, using OMP,
\begin{align}
{\bf c}_i = \arg\min_{\hat{\bf c}_i} \|{\bf x}_i-{\bf X}\hat{\bf c}_i\|_2,\qquad {\rm s.t.}\ \|{\bf c}_i\|_0=d\ {\rm and} \nonumber\\
c_{ij} = 0,\forall j\in \left(\{1,\cdots,N\}\backslash\mathcal{D}\right)\cup\{i\},\label{eq-OMP}
\end{align}
where $d$ denotes the iteration number of OMP.
Since the goal of OMP here is not calculating exactly the representation coefficients, but finding a few neighbors reliably, 
the iteration number $d$ is not very large, which will be specified later.
We then calculate the representation residual
\begin{equation}\label{eq-representation}
\mathbf{r}_i = \mathbf{x}_i - \mathbf{X} \mathbf{c}_i.
\end{equation}
After the sparse representation step, 
we \emph{update} ${\bf x}_i$ by adding an offset term to it and keeping it $\ell_2$ normalized
\begin{equation}\label{eq-update}
\mathbf{x}_i' = (\mathbf{x}_i + b\mathbf{r}_i) / \|\mathbf{x}_i + b\mathbf{r}_i\|_2,
\end{equation}
where $b>0$ denotes a modifier parameter.
Then, with certain probability $p$ we \emph{drop} this point from the future representation, 
i.e., removing index $i$ from the dictionary index set. 
The process above is repeated sequentially for all data points.
Finally, we apply spectral clustering to the similarity matrix, as done in SSC algorithms.
Notice that OMP-SSC is a special case of A-OMP-SSC, with modifier parameter $b=0$ and dropping probability $p=0$.
The detailed procedure of the proposed algorithm is included in Algorithm 1.

\begin{algorithm}[!t]
	\caption{The Proposed A-OMP-SSC}
	\begin{algorithmic}[1]
		\renewcommand{\algorithmicrequire}{\textbf{Input:}}
		\renewcommand{\algorithmicensure}{\textbf{Output:}}
		\REQUIRE Data points $\mathbf{X}\in \mathbb{R}^{D \times N}$, dictionary index set $\mathcal{D}=\{1,\cdots,N\}$, OMP iteration number $d$, modifier parameter $b>0$, dropping probability $p\in(0,1)$, cluster number $k$.
		\FOR{$i = 1, \cdots, N$}
			\STATE Find ${\bf c}_i$ using OMP (with $d$ iterations) by \eqref{eq-OMP}.
			\STATE Calcuate residual ${\bf r}_i$ by \eqref{eq-representation}.
		    \STATE Update data point: $\mathbf{x}_i \gets {\bf x}'_i$ by \eqref{eq-update}.
		    \STATE Update dictionary: $\mathcal{D} \gets \mathcal{D}\backslash\{i\}$ with probability $p$.
	    \ENDFOR
	    \STATE Apply spectral clustering (with number of clusters $k$ given) to similarity matrix $\mathbf{A}$.
	    \ENSURE Clustering labels.
	\end{algorithmic}
\end{algorithm}

\subsection{Discussion}
\label{sub:discussion}

\subsubsection{Sequentially updating data points}
\label{ssub:sequentially_updating_data_points}

\begin{figure}[!t]
    \centering
	    \includegraphics[width=0.35\textwidth]{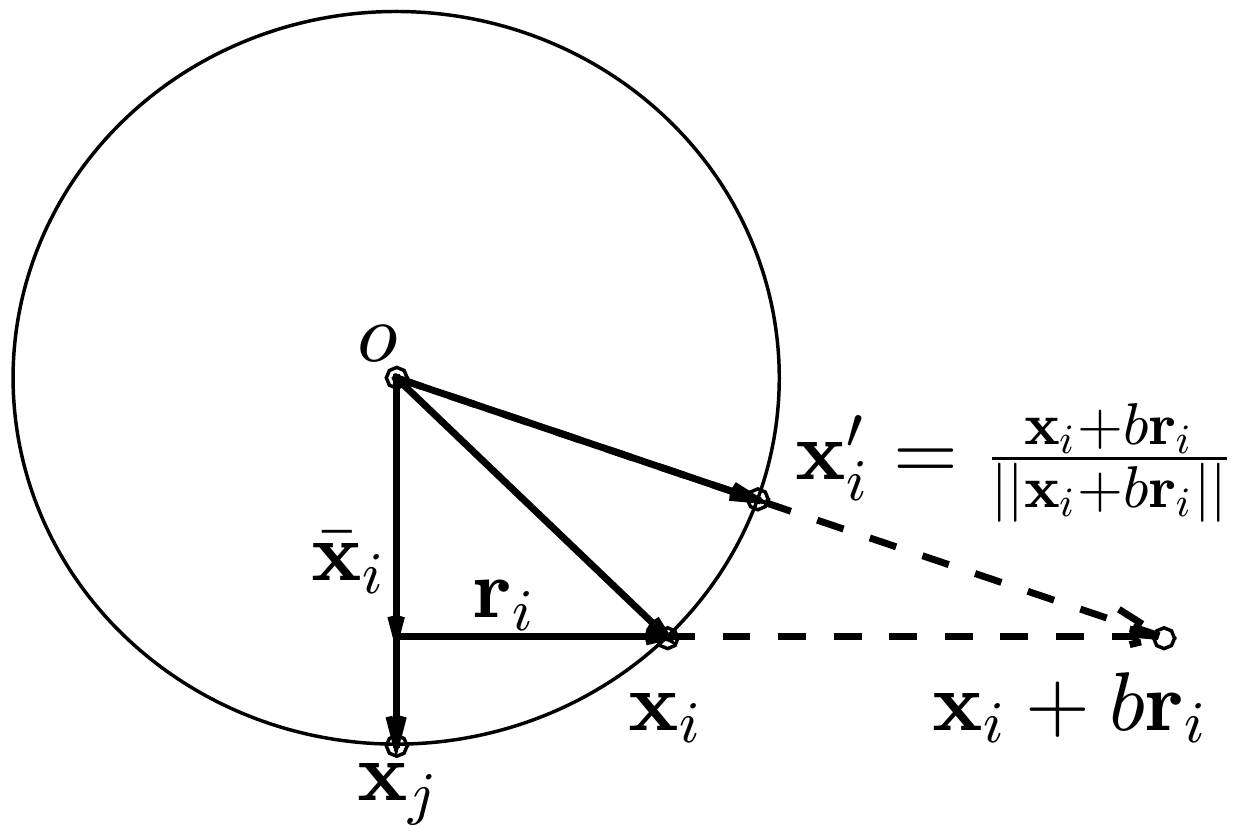}
    \caption{A visualization that updating data point $\mathbf{x}_i$ may push it further away from its neighbor ${\bf x}_j$. 
    Without loss of generality, we suppose $\mathbf{x}_i^{\rm T}\mathbf{x}_j > 0$.}
    \label{fig:active}
\end{figure}

From the construction of matrix $\mathbf{A}$ comes the intuition that, a two-direction connection doesn't contribute to connectivity.
If $\mathbf{x}_i$ chooses $\mathbf{x}_j$ ($j > i$) as its neighbor, 
we should encourage $\mathbf{x}_j$ to choose the data points other than $\mathbf{x}_i$.
By the updating operation in \eqref{eq-update}, we push $\mathbf{x}_i$ further away from its neighbors, as illustrated in Fig.~\ref{fig:active}.

This will be verified in an ideal case.
Suppose \eqref{eq-representation} is a correct decomposition of ${\bf x}_i$, 
where $\bar{\bf x}_i = \mathbf{X} \mathbf{c}_i$ lies in its ideal subspace and $\mathbf{r}_i$ in its orthogonal subspace.
Notice that the projection operation in OMP ensures that $\mathbf{x}_i^{\rm T} \bar{\mathbf{x}}_i$ is nonnegative.  
After the updating of \eqref{eq-update} we have
\begin{align}
{\mathbf{x}_i' }^{\rm T} \bar{\mathbf{x}}_i =  \frac{(\bar{\mathbf{x}}_i + (b+1)\mathbf{r}_i)^{\rm T} \bar{\mathbf{x}}_i}{\|\mathbf{x}_i + b \mathbf{r}_i\|_2} = \frac{{\bar{\mathbf{x}}_i}^{\rm T}\bar{\mathbf{x}}_i}{\|\mathbf{x}_i + b \mathbf{r}_i\|_2} 
\le \bar{\mathbf{x}}_i^{\rm T}\bar{\mathbf{x}}_i
= \mathbf{x}_i^{\rm T}\bar{\mathbf{x}}_i. \label{eq-compare}
\end{align}
The inequality in \eqref{eq-compare} comes from the fact that, when $b \ge 0$, 
\begin{align}
\|\mathbf{x}_i+b\mathbf{r}_i\|_2^2 &= \|\bar{\mathbf{x}}_i + (b+1)\mathbf{r}_i\|_2^2
= \|\bar{\mathbf{x}}_i\|_2^2 + (b+1)^2\|\mathbf{r}_i\|_2^2 \nonumber\\
&\ge \|\bar{\bf x}_i+{\bf r}_i\|_2^2 = 1, \label{norm-ge-1}
\end{align}
and the equality holds when $b=0$.

From \eqref{norm-ge-1}, $\|\mathbf{x}_i + b\mathbf{r}_i\|_2$ is symmetric about $b_0=-1$. 
In practice, since signal-to-noise ratio (SNR) of residual is generally lower than that of the data point itself, 
a positive $b$ introduces less noise than the symmetric negative one in the updating step because it has a smaller absolute value, which is preferred.
The choice of $b$ reflects a tradeoff between improving connectivity and avoiding worsening SNR too much.

\subsubsection{Randomly dropping data points}
\label{ssub:randomly_dropping_data_points}

Dropping $\mathbf{x}_i$ is more radical than updating it, as it is now impossible for $\mathbf{x}_j, j > i$ to choose $\mathbf{x}_i$ as its neighbor. 
However, dropping  data points could be risky since it reduces the data density, which is undesirable in SSC as it worsens SDP. 

\subsubsection{OMP iteration number}
\label{ssub:omp_iteration_number}

Intuitively, in OMP process, a new edge improves connectivity if it is a true connection, but weakens SDP otherwise. 
For OMP, once an atom is added to the neighbor set at a certain iteration, it cannot be removed. 
Therefore, as iteration number $d$ increases, connectivity increases but SDP decreases, as pointed out in \cite{tschannen2016noisy}.
Thus, it is not the best to choose $d$ as the subspace dimension, as done in \cite{you2016scalable}.
Choosing a smaller OMP iteration number can improve clustering accuracy and linearly decrease time consumption.

Another advantage of choosing a small OMP iteration number in our active algorithm is that, it allows more freedom for a correct choice of sparse representation coefficient satisfying SDP as $\mathbf{c}_i$ has fewer non-zero entries.
For the SC problem, the essential information from data is not the data points themselves, but their \emph{relationship} instead.
This inspires the updating and dropping steps in A-OMP-SSC, which seek to further exploit this relationship by changing the distribution of data points adaptively and thus changing the choice of representation coefficients.

\subsubsection{When it benefits}
\label{ssub:when_it_works}
A-OMP-SSC works well when data density is not too small. 
This comes from the fact that, the updating step changes the distribution of points, and the dropping step gradually decreases data density.
When the starting data density is relatively high, there are still many near neighbors for later points, so SDP would not badly decrease.

\subsubsection{Computational complexity}
\label{ssub:computational_complexity}
In OMP-SSC, each iteration requires $N$ inner products with complexity $\mathcal{O}(D)$. Thus, for $N$ points, each $d$ iterations, the total complexity of self-representation step is $\mathcal{O}(N^2 d D)$. 
For A-OMP-SSC, only an additional $\mathcal{O}(D)$ updating step is required for each point, which is neglegible.
Moreover, the dropping step gradually decreases the number of points during the process, which improves time efficiency.

\section{Numerical Experiments}
\label{sec:numerical_experiments}

In this section, we conduct numerical experiments to validate our analysis and illustrate the advantages of active OMP-SSC algorithm\footnote{The MATLAB codes for the proposed methods and all experiments are available at \url{http://gu.ee.tsinghua.edu.cn/codes/A_OMP_SSC.zip}.}.

\subsection{Synthetic Experiments}
\label{sub:synthetic_data}

We first conduct experiments to examine the effects of modifier parameter $b$, dropping probability $p$, and OMP iteration number $d$, respectively. 
Next, we compare $\ell_1$-SSC, OMP-SSC, and A-OMP-SSC algorithms, with respect to clustering error rate, connectivity, and SDP.
We also validate that A-OMP-SSC shows advantages over OMP-SSC when data density is not too small.
Finally, the superior time efficiency of the proposed algorithm is verified.
The generated synthetic samples are randomly permuted as we meet in practice.

\subsubsection{Modifier parameter $b$}
\label{ssub:residual_coefficient_c_}

\begin{figure}[tbp]
    \centering
    \begin{subfigure}{.35\textwidth}
    	\includegraphics[width = \textwidth]{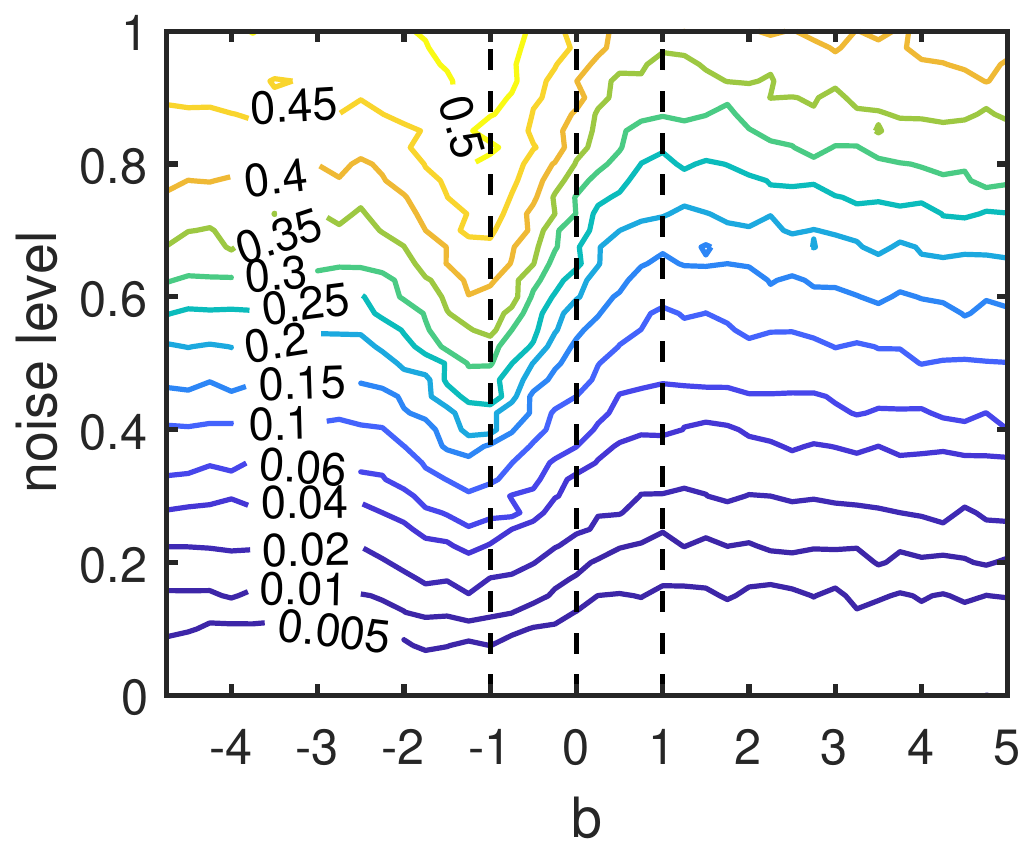}
    \end{subfigure}%
    \begin{subfigure}{.355\textwidth}
    	\includegraphics[width = \textwidth]{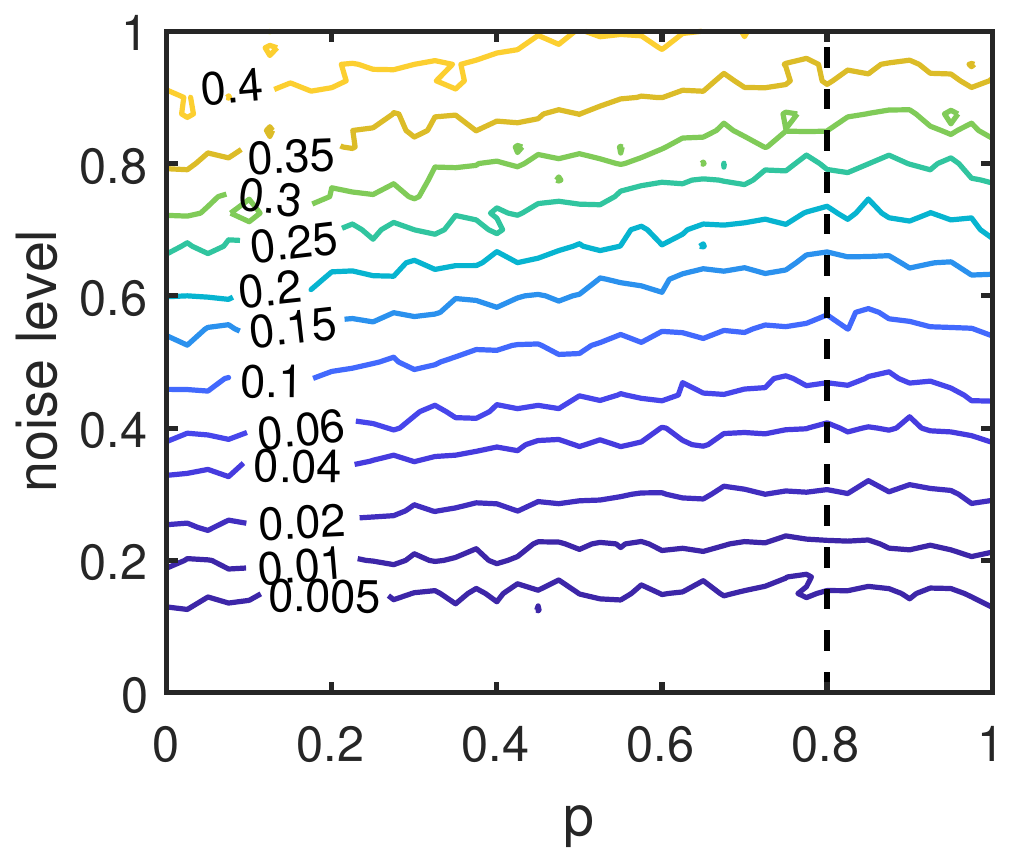}
    \end{subfigure}
    \caption{Clustering error rate versus noise level and parameter $b$ (left) or $p$ (right).}
    \label{fig:vbp}
\end{figure}

We randomly generate $3$ independent linear subspaces in $40$-dimensional ambient space, 
each of which is of dimension $6$ and has $45$ data samples. 
The noise level, i.e., additive Gaussian noise strength of each sample, varies from $0$ to $1$. 
We set $p=0$ and $d=3$.
Clustering results with different choices of modifier parameter $b$ are demonstrated in Fig.~\ref{fig:vbp} (left),
where each result is the average of $100$ independent trials.
According to Fig.~\ref{fig:vbp} (left), we read that clustering error rate is the highest at $b \approx -1$ and lowest at $1$. 
As $b$ increases from $-1$, clustering error rate first drops, leveraged by improvement in connectivity, 
but then rises due to a decrease in SNR and thus a reduction in SDP. 
(Due to limited space, results on connectivity and SDP are not shown here.)
Also notice that a positive $b$ is better than a negative one symmetric about $-1$ thanks to higher SNR.

\subsubsection{Dropping probability $p$}
\label{ssub:dropping_probability_p_}

The data is generated the same way as in the previous experiment.
We set $b=0$ and $d=3$. 
Clustering results with different choices of dropping probability $p$ are demonstrated in Fig.~\ref{fig:vbp} (right), 
which shows that clustering error rate reaches a minimum at $p \approx 0.8$.
A large $p$ up to $1$ is a good choice in this experiment, 
because the subspace dimension and OMP iteration number are small while the number of data points per subspace is relatively large.
Therefore, the decreasing data density would not be a big trouble.

\subsubsection{OMP iteration number $d$}
\label{ssub:omp_iterations}

\begin{figure}[!t]
    \centering
    \includegraphics[width = .55\textwidth]{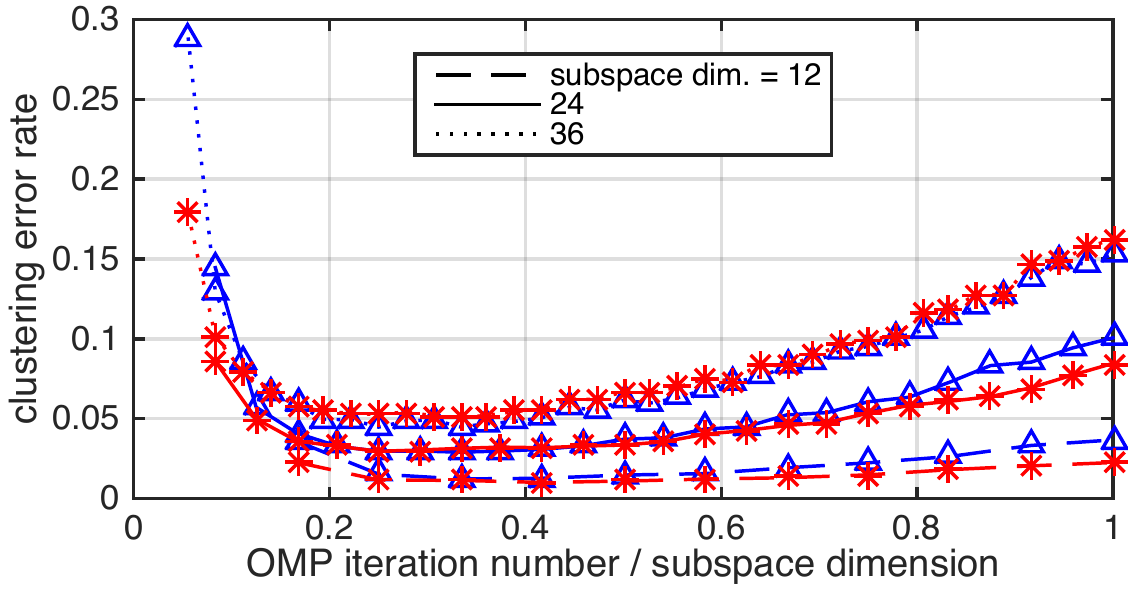}
    \caption{Effects of OMP iteration number on clustering error rate for OMP-SSC (triangle markers) and A-OMP-SSC (star markers).}
    \label{fig:varyIter_sub6_24}
\end{figure}

We randomly generate 3 independent linear subspaces in $\mathbb{R}^{120}$, each of which has $60$ data samples.
For A-OMP-SSC, $b=1$, $p=0.3$.
This experiment is conducted under three conditions, with subspace dimension set to $12$, $24$, $36$,
and the corresponding noise level $0.9$, $0.7$, $0.5$. 
Effects of OMP iteration number $d$ are demonstrated in Fig.~\ref{fig:varyIter_sub6_24}.
As $d$ increases from $1$ to subspace dimension, clustering error rate first drops owing to increasing connectivity, but then rises due to decreasing SDP, just as we discussed.
A reasonable choice of $d$ is approximately one over three of subspace dimension.
Notice that A-OMP-SSC doesn't perform well when subspace dimension is $36$, which is a result of the small data density, as we will discuss later.

\subsubsection{Performance comparison}
\label{ssub:algorithm_comparison}

\begin{figure}[!t]
	\centering
	\begin{subfigure}{.35\textwidth}
    	\includegraphics[width = \textwidth]{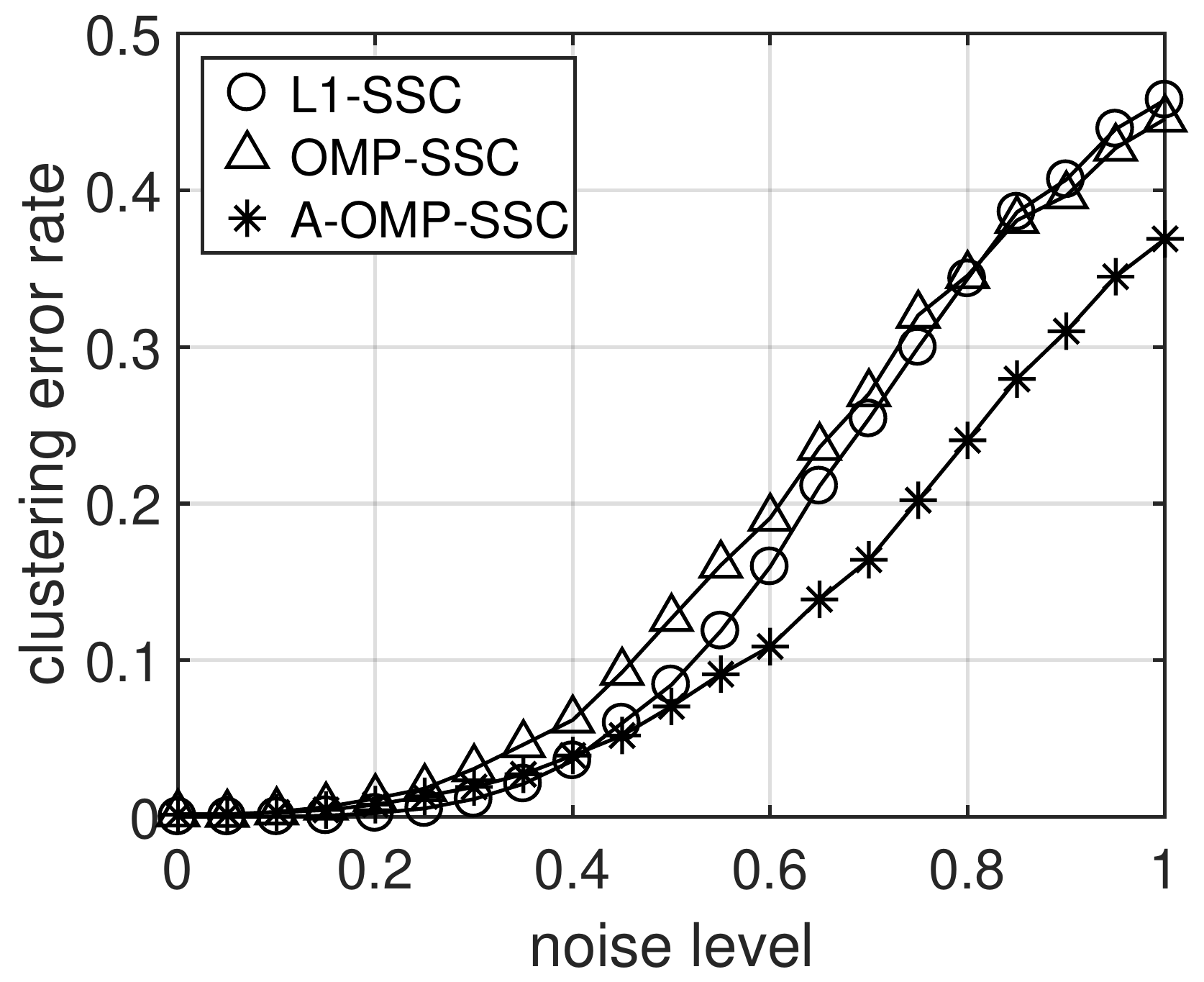}
    \end{subfigure}%
    \begin{subfigure}{.375\textwidth}
    	\includegraphics[width = \textwidth]{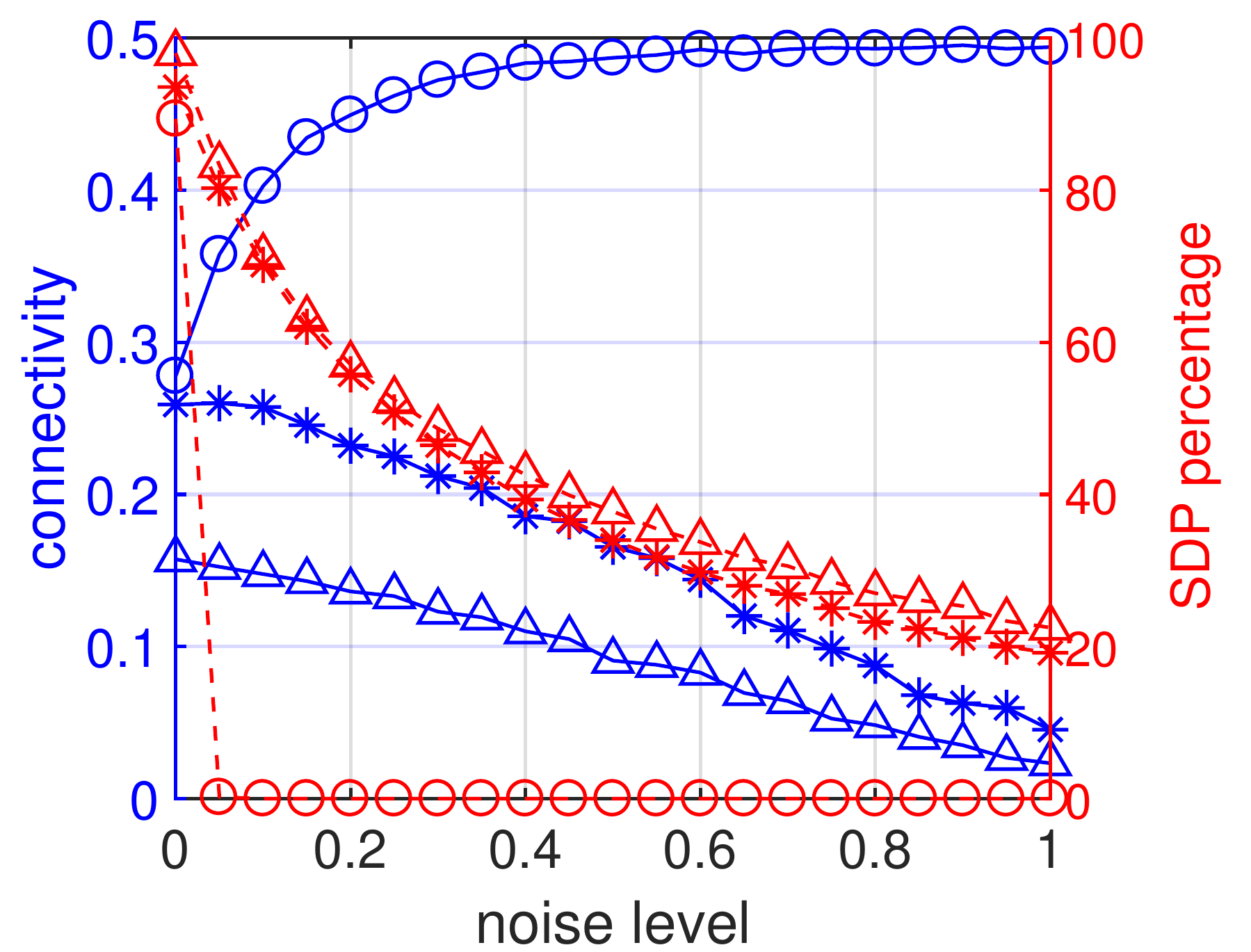}
    \end{subfigure}
	\caption{Comparison between $\ell_1$-SSC, OMP-SSC and A-OMP-SSC. Left: clustering error rate; right: connectivity (left axis, blue solid lines) and SDP percentages (right axis, red dashed lines).} 
	\label{fig:mainRst}
\end{figure}

We take the second smallest eigenvalue of the normalized Laplacian of a cluster as a metric of connectivity (with range $[0, 1]$); the larger it is, the stronger connectivity is \cite{you2016scalable}.
We also use the percentage of points satisfying SDP as a metric of SDP (with range $[0, 100]$); the larger it is, the stronger SDP is.
The data is generated the same way as in the first experiment.
Based on previous results, we set $b=1$, $p=0.8$, and $d=3$.
A performance comparison of three algorithms is demonstrated in Fig.~\ref{fig:mainRst}. 
Results show that A-OMP-SSC outperforms OMP-SSC in clustering accuracy, thanks to great improvement in connectivity, despite minor loss of SDP. 
A-OMP-SSC also outperforms $\ell_1$-SSC at slightly high noise level. This is because the number of chosen neighbors in the solution of $\ell_1$-SSC is generally around or larger than subspace dimension, therefore SDP percentage quickly decreases to zero as noise level increases.
This result demonstrates the robustness of A-OMP-SSC to white Gaussian noise.

\subsubsection{Data density} 
\label{ssub:data_density}

\begin{figure}[!t]
	\centering
	\begin{subfigure}{.35\textwidth}
    	\includegraphics[width = \textwidth]{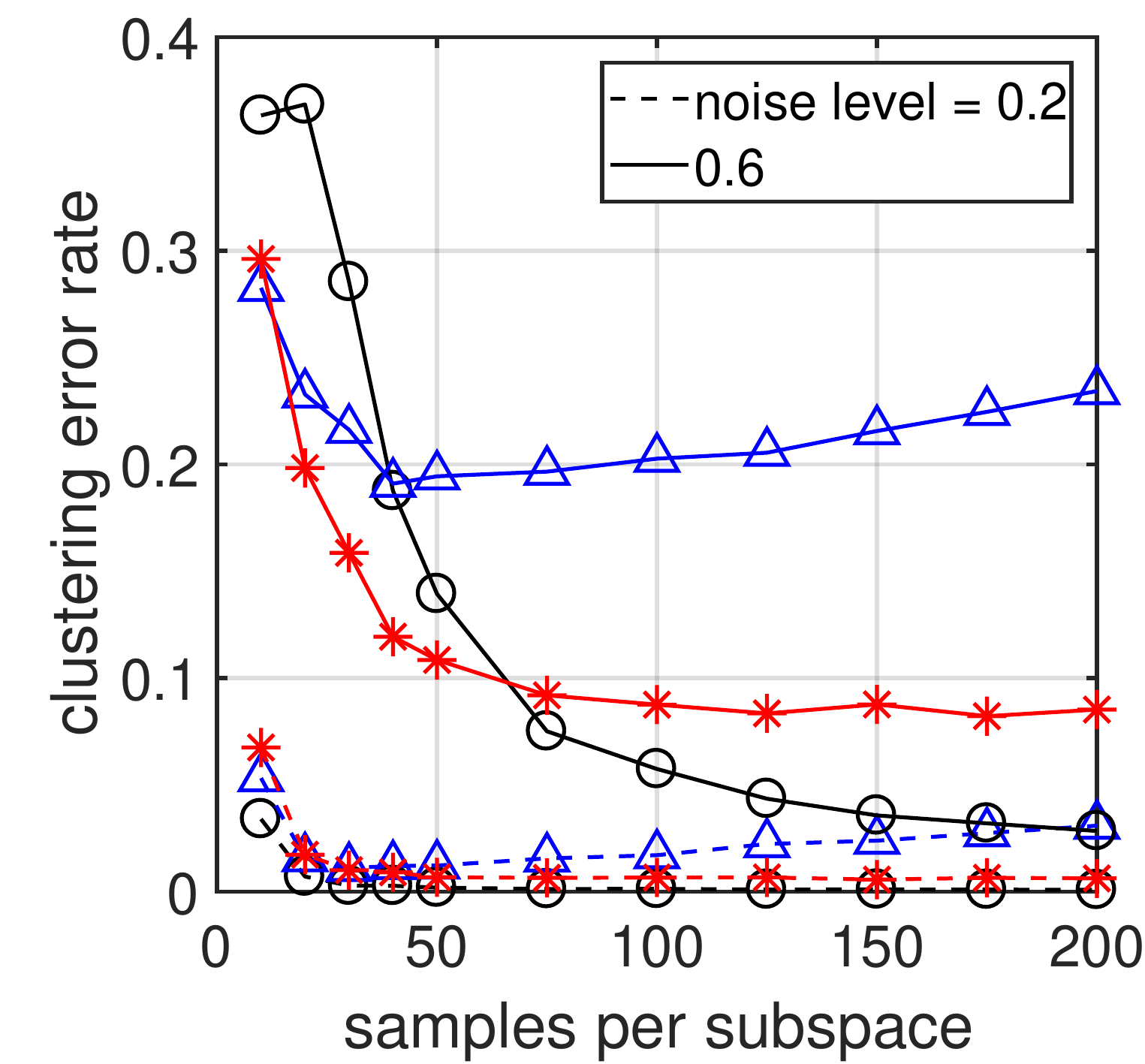}
    \end{subfigure}%
    \begin{subfigure}{.35\textwidth}
    	\includegraphics[width = \textwidth]{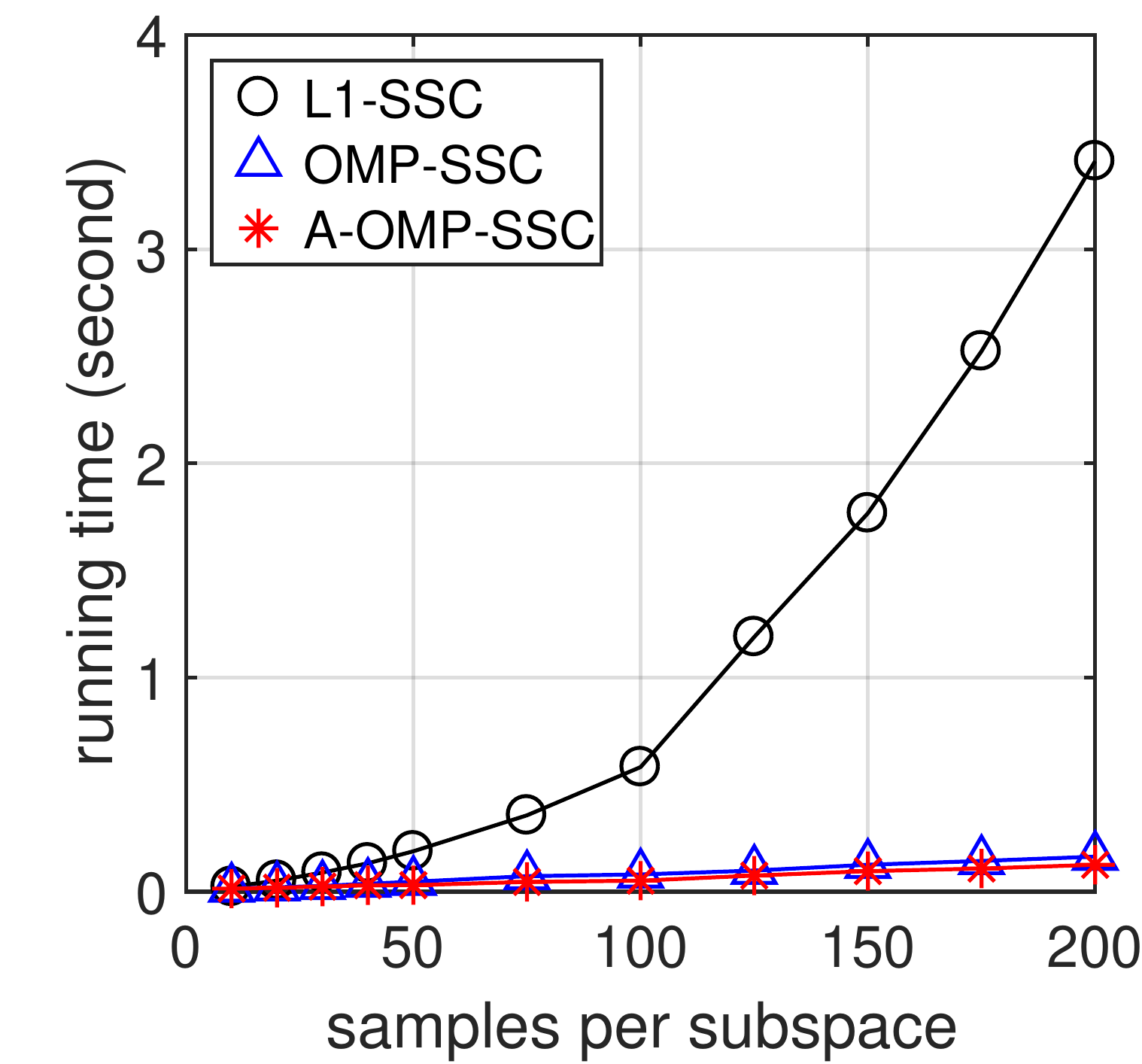}
    \end{subfigure}
	\caption{Algorithm comparisons under various problem scales. Left: clustering error rate; right: running time (in seconds).}
	\label{fig:rho_time}
\end{figure}

We use the same parameter set as the performance comparison experiment, 
except that the number of samples per subspace varies from $10$ to $200$. 
A comparison of clustering error rate between SSC algorithms under certain noise levels is demonstrated in Fig.~\ref{fig:rho_time} (left). 
OMP-SSC performs better when data density is small, while A-OMP-SSC possesses an increasing advantage as data density rises, which validates our previous analysis on the success condition of A-OMP-SSC. 

We also compare the running time of SSC algorithms under the same simulation setting.
The result is shown in Fig.~\ref{fig:rho_time} (right).
Despite the disadvantage in clustering accuracy in some scenarios, a major advantage of OMP-SSC is that it is much faster than traditional $\ell_1$-SSC, which is consistent with the results from existing literature \cite{you2016scalable,tschannen2016noisy}.
Furthermore, A-OMP-SSC is even faster than OMP-SSC, because the operation of updating data points doesn't change the overall computational complexity, while the operation of dropping data points gradually reduces data density.

\subsection{Face Clustering}
\label{sub:face_clustering}

We test our algorithm on EYaleB, a dataset of human-face pictures from 38 persons, each with around 64 photos under various illumination. 
In our experiment, the pictures are downsampled to $48 \times 42$. 
It is known that human-face images of a subject can be approximated by a 9-dimensional subspace \cite{elhamifar2013sparse}.
We set $b=0.5$, $p=0.2$ and $d=3$.
A comparison of OMP-SSC and A-OMP-SSC on clustering error rate is demonstrated in Table~\ref{tab:yale2016}.
The result is the average of $200$ trials. 
For each trial, given the number of clusters $k$, we randomly choose $k$ subjects, mix all images of these subjects together, and permutate them before applying clustering algorithms. 
The result shows that A-OMP-SSC outperforms its reference on human face clustering problem,
especially in the scenario of multiple classes.
Notice that the state-of-the-art clustering error rate of $\ell_1$-SSC in the case of all $38$ subjects is $31.0\%$ \cite{you2016scalable}, while its running time is prohibitive. 
This highlights the fact that A-OMP-SSC enjoys both superior time efficiency and higher clustering accuracy in such large-scale problems.

\begin{table}[!t]
    \centering
    \small
    \caption{EYaleB clustering error rate in percentage.}
    \begin{tabular}{c|c|c|c|c|c|c|c}
    \hline
    \# clusters & 2 & 3 & 5 & 8 & 13 & 21 & 38 \\
    \hline
    OMP-SSC & 1.11 & 2.64  & 5.52  & 8.43  & 14.4 & 19.8 & 28.7 \\
    \hline
    A-OMP-SSC     & {\bf 1.10} & {\bf 2.23} & {\bf 4.11} & {\bf 6.94} & {\bf 12.2} & {\bf 16.8} & {\bf 22.3} \\
    \hline
    \end{tabular}
    \label{tab:yale2016}
\end{table}

\section{Conclusion}
\label{sec:conclusion}

We propose a fast and accurate A-OMP-SSC algorithm, which obtains a better tradeoff between connectivity and SDP via adaptively updating and randomly dropping data points.
We heuristically explain the intuitions behind our algorithm and analyze the condition where it works.
Numerical results validate our analyses, and demonstrate the advantages of our algorithm over OMP-SSC and $\ell_1$-SSC.
The proposed active mechanism may benefit many other SC algorithms.

\bibliographystyle{ieeetr}
\bibliography{refs_SSC}

\end{document}